\documentclass{article}
\usepackage{spconf,amsmath,graphicx}

\usepackage{verbatim}
\usepackage{multirow}
\usepackage{hyperref}
\usepackage{color}
\usepackage{mathtools}
\usepackage{pifont}

\title{Latent Dirichlet Allocation Based Organisation of Broadcast Media Archives for Deep Neural Network Adaptation}
\name{Mortaza Doulaty, Oscar Saz, Raymond W. M. Ng, Thomas Hain
	\thanks{This work was supported by the EPSRC Programme Grant EP/I031022/1 (Natural Speech Technology).}
}
\address{\normalsize{Speech and Hearing Group, Department of Computer Science, University of Sheffield, Sheffield, S1 4DP, UK}}
\begin{document}
%
\maketitle
\begin{abstract}
This paper presents a new method for the discovery of  latent domains in diverse speech data, for 
the use of adaptation of Deep Neural Networks (DNNs) for Automatic Speech Recognition. Our 
work focuses on transcription of multi-genre broadcast media, which is often only categorised 
broadly in terms of high level genres such as sports, news, documentary, etc. However, in terms of 
acoustic modelling these categories are coarse. Instead, it is expected that a mixture of latent 
domains can better represent the complex and diverse behaviours within a TV show, and therefore lead 
to better and more robust performance. We propose a new method, whereby these latent domains are 
discovered with Latent Dirichlet Allocation, in an unsupervised manner. These are used to adapt DNNs 
using the Unique Binary Code (UBIC) representation for the LDA domains. Experiments conducted on a 
set of BBC TV broadcasts, with more than 2,000 shows for training and 47 shows for testing, show 
that the use of LDA-UBIC DNNs reduces the error up to 13\% relative compared to the baseline hybrid 
DNN models.
	


\end{abstract}
%

\section{Introduction}
\label{sec:intro}
Streaming and webcasts are popular in this age of high--speed internet and mobile networks.
With the ever increasing amount of audio--visual media data, the ability to index their 
contents and search for them is becoming more and more important. For data with speech contents, 
using Automatic Speech Recognition (ASR) to get the transcripts, is an efficient way to search and 
browse through thousands of hours of recordings. Error rates for the traditional broadcast news 
programmes could reach below 10\% even in 1990s \cite{Woodland97, Gauvain02, Gales06}. 
However broadcast media is not just limited to clean and read studio speech but also includes other 
types of multi--genre data with diverse speakers, variety of acoustic and recording conditions and 
diversity of the topics covered resulting in complex acoustic, lexical and linguistic conditions 
which is not yet well studied \cite{Lanchantin13}.

The wide variety of conditions in complex broadcast media causes mismatch between training and 
testing data, and therefore degrades the performance of the speech recognition systems \cite{doulatyis15}. Adaptation 
can compensate for this mismatch. For Gaussian Mixture Model/Hidden Markov Model (GMM/HMM) systems 
several well established adaptation methods exist. However, adaptation of Deep Neural Networks 
(DNNs) is still a very active research topic. DNN adaptation methods can be divided into these three 
main categories \cite{deng2015}: 

\begin{enumerate}
	\item  Linear input transformations: this is the most common adaptation method where 
a linear transformation is applied to either input feature \cite{Abrash95connectionistspeaker}, 
input to the softmax layer \cite{li2010comparison} or activation of the hidden layers 
\cite{gemello2007linear}
	
	\item Retraining: all or some of the model parameters are adapted or trained using the 
adaptation data \cite{stadermann2005two,doddlipatla_is14}.
	
	\item Subspace methods: a speaker/environment subspace is estimated and then neurons' 
weights or transformations are computed, based on the subspace representation of the 
speaker/environment. Principle Component Analysis (PCA) based adaptation approach 
\cite{dupont2000fast}, i-Vector based speaker--aware training \cite{saon2013speaker} 
or speaker--aware DNNs \cite{liu2015} 
can be considered as subspace methods.
\end{enumerate}

Broadcast media is complex in nature. For instance, in a news programme, there are in--studio 
reporting and live coverage on the scene. Assuming that all content variation from a single show 
can be described by a single, vaguely--defined domain is unrealistic and also not that helpful to 
ASR. Nonetheless it is clear that certain show types have very specific characteristics. 
Being able to assign broadcast media to a mixture of domains can alleviate this problem. Latent 
Dirichlet Allocation (LDA) is a statistical approach to discover latent variables in a collection
of data that is describable with first--order statistic, in an unsupervised manner 
\cite{blei2003latent}. It is mostly used in Natural Language Processing (NLP) for the categorisation 
of text documents, but it has been used for audio and image processing as well. In audio tasks, LDA 
has been used for classifying unstructured audio files into onomatopoeic and semantic descriptions 
with successful results \cite{kim2009acoustic}. We have previously used LDA for domain adaptation of 
GMM/HMM systems \cite{doulaty2015lda}. 

This paper builds on this knowledge, and introduces a method on how to use LDA for domain 
adaptation of hybrid DNNs. Using LDA models, a data class - 
further referred to as ``LDA domain'',  is chosen for each utterance. The class information is then
provided to the DNN in training. The learning algorithm adjusts the model parameters to 
exploit these additional information. During testing the same information is supplied. 
We further refer to this method as Latent--Domain--aware Training (LDaT). Results shown
later in this paper indicate significant improvements of LDaT over baseline and input--adapted DNNs.

The following section briefly introduces LDA, followed by a description of acoustic data LDA. In 
section \ref{sec:lda-dnn}, DNN adaptation using LDA is described. Section \ref{sec:exp} describes 
the experimental setup, followed by discussion and a conclusion.

\section{Latent Dirichlet Allocation}
\label{sec:lda}
Latent Dirichlet Allocation (LDA) \cite{blei2003latent} is an unsupervised probabilistic generative 
model for collections of discrete data. It aims to describe how every item within a collection is 
generated, assuming that there are a set of latent variables and that each item is modelled as a 
finite mixture over those latent variables. 
LDA was originally used for topic modelling of text corpora; however, it is a generic model and can 
be applied to other tasks, such as object categorisation and localisation in image processing 
\cite{sivic2005discovering}, automatic harmonic analysis in music processing 
\cite{hu2009probabilistic}, acoustic information retrieval in unstructured audio analysis 
\cite{kim2009acoustic} and our previous work for domain adaptation of GMM/HMM systems 
\cite{doulaty2015lda}.

A dataset is defined as a collection of sets where each set is in turn a collection of discrete 
symbols (in case of topic modelling of text documents, a document is equivalent to a set and words 
inside a document are equivalent to the discrete symbols).
Each set is represented by a $V$-dimensional vector based on the histogram of the symbols' table 
which has size of $V$.
It is assumed that the sets were generated by the following generative process:
\begin{enumerate}
	\item For each set $d_m, m \in \{1 ... M\}$, choose a $K$--dimensional latent variable weight vector $\theta_m$ from the Dirichlet distribution with scaling parameter $\alpha$: $p(\theta_{m}|\alpha)=Dir(\alpha) $
	\item For each discrete item $w_n, n \in \{1 ... N\}$ in set $d_m$ 
	\begin{enumerate}
		\item Draw a latent variable $z_n \in \{1 ... K \}$ from the multinomial distribution $p(z_n=k|\theta_m)$ 
		\item Given the latent variable, draw a symbol from $p(w_n | z_n, \beta)$, where $\beta$ is a $V \times K$ matrix and \\$\beta_{ij}=p(w_n = i | z_n = j, \beta)$
	\end{enumerate}
\end{enumerate}
It is assumed that each set can be represented as a bag--of--symbols - i.e. by first--order 
statistics, which means any symbol sequence relationship is disregarded. Since speech and text 
are highly ordered processes this can be an issue. Another assumption
is that the dimensionality of the Dirichlet distribution  $K$ is fixed and known (and thus the 
dimensionality of the latent variable $z$). 

\begin{figure}
	\centering
	\includegraphics[width=6cm]{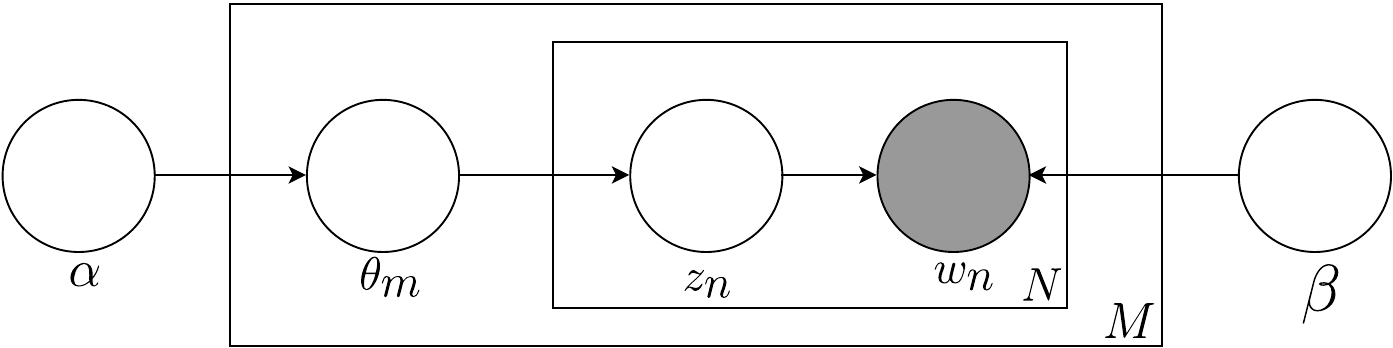}
	\caption{Graphical model representation of LDA}
	\label{fig:lda-graphical-model}
\end{figure}

A graphical representation of the LDA model is shown at Figure \ref{fig:lda-graphical-model} as a 
three--level hierarchical Bayesian model. In this model, the only observed variable is $w$ and the 
rest are all latent. $\alpha$ and $\beta$ are dataset level parameters, $\theta_m$ is a set level 
variable and $z_n$, $w_n$ are symbol level variables. The generative process is described formally 
as:
\vspace{-1mm}
\begin{equation}
p(\theta, \mathbf{z}, \mathbf{w} | \alpha, \beta) 
= p(\theta | \alpha) \prod_{n=1}^{N}p(z_n | \theta) p(w_n|z_n,\beta)
\end{equation}
\vspace{-1mm}
The posterior distribution of the latent variables given the symbols and $\alpha$ and $\beta$ parameters is:
\vspace{-1mm}
\begin{equation}
\label{eq:posterir}
p(\theta, \mathbf{z} | \mathbf{w}, \alpha, \beta) = 
\frac{p(\theta, \mathbf{z}, \mathbf{w} | \alpha, \beta)}{p(\mathbf{w} | \alpha, \beta)}
\end{equation}
Computing $p(\mathbf{w} | \alpha, \beta)$ requires some intractable integrals. A reasonable approximate can be acquired using variational approximation, which is shown to work reasonably well in various applications \cite{blei2003latent}. The approximated posterior distribution is:
\vspace{-1mm}
\begin{equation}
\label{eq:approx-posterior}
q(\theta, \mathbf{z} | \gamma, \phi) = q (\theta | \gamma) \prod_{n=1}^{N}q(z_n | \phi_n)
\end{equation}
where $\gamma$ is the Dirichlet parameter that determines $\theta$ and $\phi$ is the parameter for the multinomial that generates the latent variables. 

Training tries to minimise the Kullback--Leiber Divergence (KLD) between the real and the 
approximated joint probabilities (equations \ref{eq:posterir} and \ref{eq:approx-posterior}) 
\cite{blei2003latent}: 
\vspace{-1mm}
\begin{equation}
\underset{\gamma, \phi}{argmin} 
\; KLD \big(
q(\theta, \mathbf{z} | \gamma, \phi)
\; || \; 
p(\theta, \mathbf{z} | \mathbf{w}, \alpha, \beta)
\big)
\end{equation}
\vspace{-1mm}
Other training methods based on Markov--Chain Monte-Carlo are also proposed, like Gibbs sampling 
method \cite{griffiths2004finding}.

\section{Acoustic LDA}
\label{sec:lda-adapt}
As outlined above, LDA is a model to describe latent factors in sets of discrete 
symbols \cite{blei2003latent} which are here interpreted as ``domains''. In order to  fit into that 
concept speech signals need to be converted into such a form. Typically speech is represented using 
continuous features (e.g. with Mel frequency cepstral coefficients), and has variable length. In our 
previous work \cite{doulaty2015lda} we used Linde--Buzo--Gray vector quantization algorithm \cite{gersho1992vector} to represent each speech 
frame with a discrete symbol, equivalent to an acoustic word or phone label. 

In this paper an approach similar to that used in \cite{ni2015dataselection} was implemented. A 
GMM model with $V$ components is trained using all of the training data. The model is then used to 
get the posterior probabilities of the Gaussian components to represent each frame with index of the 
Gaussian component with the highest posterior probability. Frames of every speech segment of length 
$T$, $\mathbf{x}=\{\mathbf{x}_1,...,\mathbf{x}_t,...,\mathbf{x}_T\}$ are represented as:
\begin{equation}
\tilde{x}_t=\underset{i}{argmax} \; P(G_i | \mathbf{x}_t) \
\end{equation}
where $G_i \; (1 \le i \le V)$ is the $ith$ Gaussian component. After applying this 
process to each utterance, each speech segment is represented as $\{ 
\tilde{x}_1,...,\tilde{x}_t,...,\tilde{x}_T \}$ where $x_t$ is index of the Gaussian component and thus a natural number $(1 \le x_t \le V)$. Here we refer to each speech utterance as an acoustic document.
With this information, a fixed length vector $\mathbf{\hat{x}} = \{ a_1,...,a_i,...,a_V \}$ of size $V$ were constructed to represent the count of every Gaussian component in an acoustic document. 
%
This leads to a type of bag--of--sounds representation. The sounds would normally be expected to relate
to phones, however given the acoustic diversity of background conditions many other factors may 
play a role. Once these bag--of--sounds representations of acoustic documents are derived, LDA models can be trained. 

\section{LDA--DNN Adaptation}
\label{sec:lda-dnn}
After acoustic symbols are established and speech segments are represented as bag--of--sounds, LDA 
models with designated latent domain sizes
are trained using the variational EM algorithm \cite{blei2003latent}. Hence, the posterior 
distribution of latent domains ($z_m$) for each utterance $m$ is computed. 
Since there can be many utterances in the training set, to effectively incorporate domain 
information in the vast amount of data, each utterance is assigned to only one domain. The 
assignment is made according to the maximum posterior estimate of domains $p(z_m)$. 

The maximum posterior assumption requires high domain homogeneity for each acoustic 
document. This can to some degree be controlled by the size of domains. With a large number of 
domains, the resolution may be too high and the domain homogeneity within one acoustic document may 
be therefore lowered. On the other hand it is desirable to have a sufficient number of domains such 
that the variability in shows and between different types of shows are sufficiently covered.  


Finally, domain information derived from the LDA model with $K$ domains is encoded with a 
$K$-dimensional one--hot vector called Unique Binary Index Code (UBIC) \cite{liu2015}. UBIC 
indicates the most likely domain of the utterance using the posterior domain probability. UBIC is 
then used to augment the input feature vectors. Apart from the extra nodes and connections in the 
input layer, the DNN architecture is identical to other baseline DNNs which are not domain--aware. 

\begin{figure}
	\centering
	\includegraphics[width=8cm]{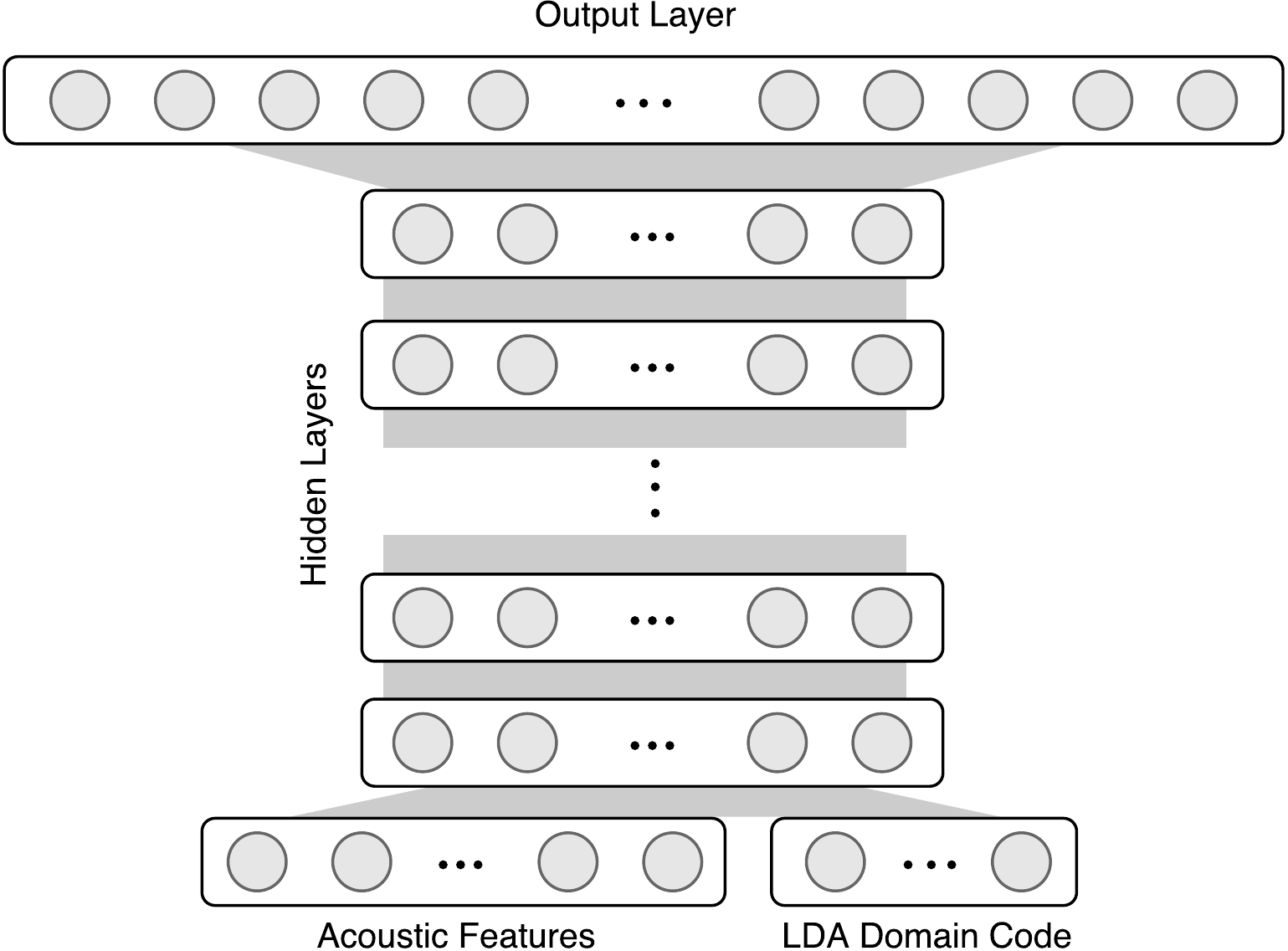}
	\caption{LDA-DNN Topology}
	\label{fig:lda-dnn-topo}
\end{figure}

With the baseline DNNs, activation of the first layer is:
\begin{equation}
\label{eq:dnn}
\mathbf{v}^1 = f(\mathbf{W}^1 \mathbf{v}^0 + \mathbf{b}^1)
\end{equation}
where superscripts denote the layer index, $\mathbf{v}^1$ is the activation vector of the first layer, $\mathbf{W}^i$ and $\mathbf{b}^i$ are the weight matrix and bias vector associated with layer $i$ and $\mathbf{v}^0$ is the input features. With augmented UBIC in LDaT training this becomes:
\begin{equation}
\label{eq:dnn-lda}
\begin{aligned}
\mathbf{v}^1_{LDaT} & =  f \Big( \begin{bmatrix}  \mathbf{W}^1_\mathbf{v}  \mathbf{W}^1_\mathbf{d} \end{bmatrix}  \begin{bmatrix} \mathbf{v}^0\\\mathbf{d} \end{bmatrix} + \mathbf{b}^1_{LDaT}  \Big) \\
& = f \Big( \mathbf{W}_v^1 \mathbf{v}^0  + \underbrace{\mathbf{W}_d^1 \mathbf{d} +  \mathbf{b}^1_{LDaT}}_\text{domain specific bias}  \Big) \\
\end{aligned}
\end{equation}
where $\mathbf{d}$ is the $K$--dimensional domain assignment vector from the LDA model, $\mathbf{W}^1_v$ is the weigh matrix for the acoustic features and it is initialised from $\mathbf{W}^1$ of equation \ref{eq:dnn}. $\mathbf{W}^1_s$ is the weigh matrix for the augmented LDA domain assignment input. Comparing equations \ref{eq:dnn} and \ref{eq:dnn-lda} the only difference is in the bias vector where there was a fixed bias before ($\mathbf{b}^1$) and now with the augmented LDA domain information, there is a new adapted bias $\mathbf{b}_d^1 = \mathbf{W}_d^1 \mathbf{d} +  \mathbf{b}^1_{LDaT}$ for each of the LDA domains. This type of adaptation is efficient, since it is implicit in the training process and does not require further adaptation steps \cite{deng2015}.
Figure \ref{fig:lda-dnn-topo} illustrates the DNN architecture with the augmented UBIC code.

\section{Experimental Setup}
\label{sec:exp}
\subsection{Data}
TV broadcasts from the BBC were selected for the experiments. The data is identical to the one 
defined and provided for the 2015 Multi--Genre Broadcast (MGB) Challenge \cite{MGB, MGB_ASR_SHEF,ng2014iwslt}. The shows 
were chosen to cover the full range of broadcast show types in current TV, and categorised in 
terms of 8 genres: advice, children's, comedy, competition, documentary, drama, events and news. 
Acoustic Model training data was fixed and limited to more than 2,000 shows, broadcast by the 
BBC during 6 weeks in April and May of 2008. The development data for the task was 47 shows 
broadcast by the BBC during a week in mid--May 2008. The amount of shows and broadcast time for 
training and development data is shown in Table \ref{tab:mgbdata}.

\begin{table} [th]
	\caption{\label{tab:mgbdata} {Amount of training and development data}}
	\centerline{
		\begin{tabular}{|c|c|c|c|c|}
			\hline
			&  \multicolumn{2}{|c|}{Train} & \multicolumn{2}{|c|}{Development} \\
			\hline 
			Genre & Shows & Time & Shows & Time \\
			\hline
			Advice & 264 & 193.1h & 4 & 3.0h \\
			Children's & 415 & 168.6h & 8 & 3.0h \\
			Comedy & 148 & 74.0h & 6 & 3.2h \\
			Competition & 270 & 186.3h & 6 & 3.3h \\
			Documentary & 285 & 214.2h & 9 & 6.8h \\
			Drama & 145 & 107.9h & 4 & 2.7h \\
			Events & 179 & 282.0h & 5 & 4.3h \\
			News & 487 & 354.4h & 5 & 2.0h \\
			\hline
			Total & 2,193 & 1580.5h  & 47 & 28.3h  \\
			\hline
		\end{tabular}}
	\end{table}

For the training data high quality transcription was not available. Instead only the subtitle 
text broadcast with each show plus an aligned version of the subtitles were available where the 
time stamps of the subtitles had been corrected in a lightly supervised manner \cite{Long13}. After 
this process, the new transcripts for the training shows had two potential problems: first, the 
subtitle text might not always match the actual spoken words and second, the time boundaries given 
might have errors arising from the lightly supervised alignment. To alleviate these two problems, 
only segments with Word Matching Error Rate (WMER) of lower than $40\%$ were used, which yielded 
around 500h of data. The WMER was a by--product of the semi--supervised alignment process that 
measures how similar the text in the subtitle matched the output of a lightly supervised ASR system 
for that segment \cite{Long13}.

For the Language Model (LM) subtitles from shows broadcast from 1979 to March 2008, with a total 
of 650 million words were used to train statistical language models. 

\subsection{Baseline}

Initial models were GMM/HMM systems with 13 dimensional PLP \cite{hermansky1990perceptual} features 
where four neighbouring frames on each side were spliced together to form a 117--dimensional feature 
vector. Using Linear Discriminant Analysis \cite{haeb1992linear} this feature vector was projected 
down to 40--dimensional vector and a global Constrained Maximum Likelihood Linear Regression 
\cite{gales1998mllr} transformation was applied to de--correlate the features. Speaker Adaptive 
Training (SAT) \cite{anastasakos1996compact} was performed and then models were 
discriminatively trained using the Boosted Maximum Mutual Information criterion \cite{povey2008boosted} 
and used to get the state level alignments for the DNN training. The input to the DNN was 440 
dimensional PLP features that were $\pm$5 frames to the left/right of the current frame. The network 
had 6 hidden layers of size 2048 and an output layer of size 6478. The network was initialised using 
Deep Belief Network \cite{hinton2006fast} pre--training and then trained to optimise per 
frame Cross Entropy objective function with Stochastic Gradient Descent. A speaker 
adapted DNN was also trained as the second base--line system using SAT style training. 
Speaker--based CMLLR transformations were applied to the features to make the inputs of the DNN 
closer to an average speaker. The Kaldi open--source speech recognition toolkit \cite{povey_2011_kaldi} 
was used to train the acoustic models.


For decoding a 50k lexicon with a highly pruned 3--gram language model was used to generate lattices and then those lattices were re--scored using a 4--gram language model. Both of the language models were trained on the 650M words of the subtitles data using the SRILM toolkit \cite{stolcke2002srilm}

Table \ref{tab:baseline} presents the Word Error Rate (WER) of the development set with baseline models. There is a 19\% relative WER reduction from GMM/HMM models to the baseline DNN models as usually expected. Speaker--adapted DNN also yields a further 6\% relative WER reduction compared to the un--adapted DNN.

\begin{table}[]
	\centering
	\caption{Baseline}
	\label{tab:baseline}
	\begin{tabular}{|c|c|c|}
		\hline
		\multicolumn{2}{|c|}{Model}     & WER (\%)  \\ \hline
		GMM                  & SAT BMMI & 41.0 \\ \hline
		\multirow{2}{*}{DNN} & Baseline & 33.3 \\ \cline{2-3} 
		& Speaker Adapted  & 31.4 \\ \hline
	\end{tabular}
\end{table}

\subsection{LDA--DNN Experiments}
A GMM with 4k Gaussian mixtures is constructed using the mix--up procedure. Using this GMM, the audio frames are mapped to discrete symbols to train the LDA models \cite{rehurek_lrec}. With LDA models, we experimented with different number of latent domains, namely $4$, $6$, $8$, $16$, $32$, $64$, $128$, $256$ and $512$.

For each of the domain sizes mentioned above, we computed the average domain entropy over all acoustic documents.
Entropy increases from $0.76$ to $4.06$ when domain size increases from $4$ to $512$. 
In this experiment, domain sizes $64$ and $128$ were used. This leveraged the considerations about the homogeneity and sparsity of the discovered domains discussed in section \ref{sec:lda-dnn}.

\begin{figure}
	\centering
	\includegraphics[width=8.5cm]{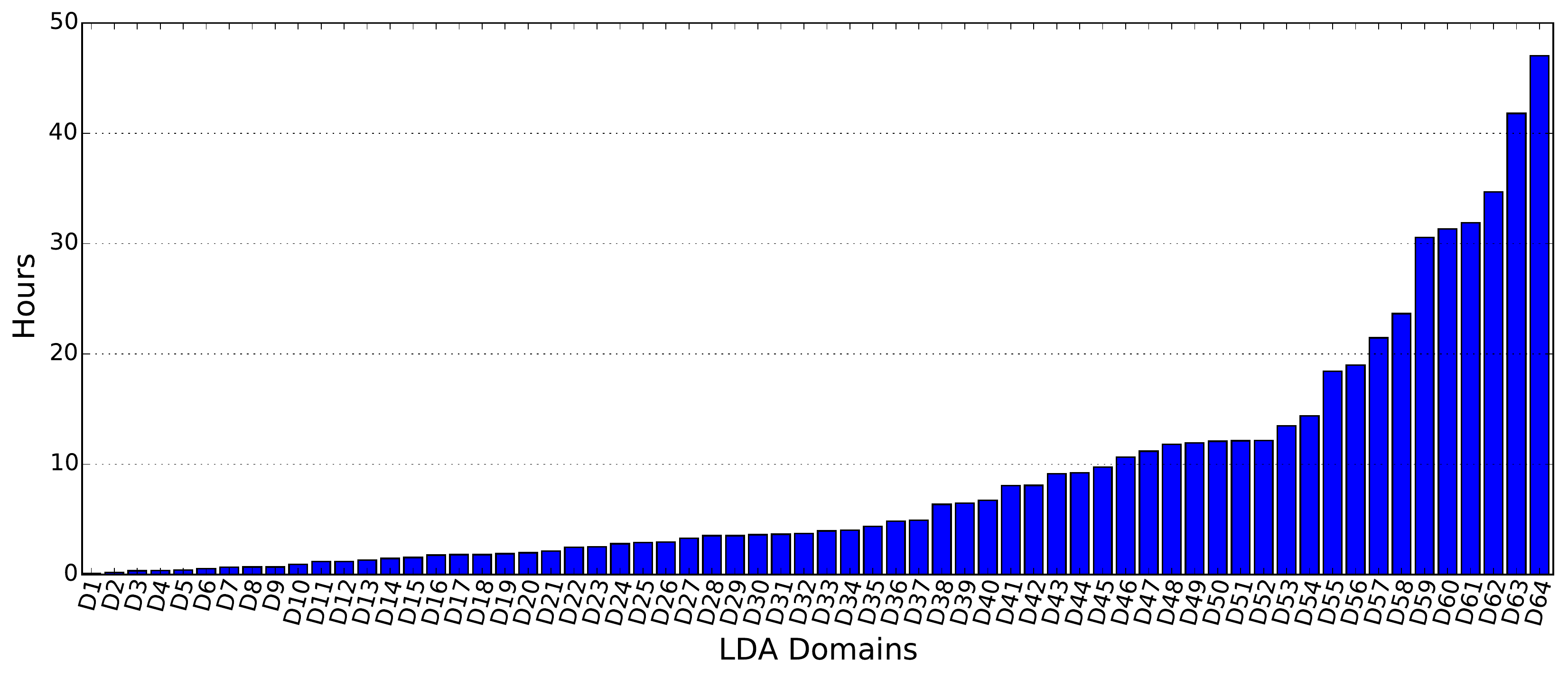}
	\caption{Distribution of data across LDA domains}
	\label{fig:lda-data-dist}
\end{figure}

Apart from selecting an appropriate size of domain, cross-agreement data filtering
was performed to ensure high domain homogeneity for each acoustic document. A domain-tuple with $8192$ items was established. These items come from the Cartesian product of the $64 \times 128$ domain mappings from the two corresponding LDA models. It is assumed that the two LDA models share a significant portion of the domains. If there is a high heterogeneity within an acoustic document, maximum-a-posteriori domain assignment from either or both LDA models will not be accurate, and they would appear in the rare classes in the $8192$ domain--tuple items. Histogram pruning based on normalised pairs counts was performed to remove those rare items. The pruning cut--off was determined to result in a target training set size of around 500h, which was comparable to the data amount in our previous baseline experiments. Figure \ref{fig:lda-data-dist} shows the amount of data (in hours) for each of the 64 LDA domains. 

The baseline DNN systems had an input layer of size $440$. That input was expanded by augmenting the LDA inferred domain with one--hot encoding. The new input had the size of $504$ ($440+64$). The new LDA--DNN was trained similarly to the base line DNNs. Table \ref{tab:frame-accuracy} shows the frame classification accuracy of DNNs on a $10\%$ held--out cross--validation set with and without augmenting UBIC vectors.

\setlength\tabcolsep{3.5pt}
\begin{table}[h]
	\centering
	\caption{Frame classification accuracy with and without LDA UBIC vectors}
	\label{tab:frame-accuracy}
	\begin{tabular}{|c|c|c|}
		\hline
		\multirow{2}{*}{Model}              & \multicolumn{2}{c|}{CV Set Frame Accuracy (\%)}    \\ \cline{2-3} 
		& Without LDA UBIC      & LDA UBIC         \\ \hline
		Un--adapted DNN                            & 50                      & 50                      \\ \hline
		Speaker Adapted DNN                      & 48
		& 46                    \\ \hline
		
	\end{tabular}
\end{table}

Table \ref{tab:all-res} presents the WER of baseline and adapted models for all of the eight genres. LDaT training reduces the WER from 33.3\% to 30.6\%, which is even better than speaker adapted DNN (31.4\%). Combining speaker adaptation and domain adaptation (SAT+LDaT, linear input transformation for the speaker and bias adaptation for the latent domain) yields 28.9\%, which is 13\% relative WER reduction compared to the baseline DNN model and 8\% relative improvement over the speaker adapted DNN. This also suggests that LDA inferred domains were not speaker clusters (since combining two adaptations still improves the performance). 
Because of the diverse nature of the data used, WER differs a lot across genres. Namely comedy and drama had the highest errors (43.8\% and 45.0\% respectively with LDaT+SAT models) showing the difficult nature of these genres. On the other hand, news had the lowest WER (14.3\%). The WER diversity across the genres was consistent between all of the four models presented in table \ref{tab:all-res}.

\setlength\tabcolsep{8.5pt}
\begin{table*}[th]
	\centering
	\caption{Per--genre WER for all of the models}
	\label{tab:all-res}
\begin{tabular}{|c|c|c|c|c|c|c|c|c||c|}
	\hline
	\multirow{2}{*}{Adaptation} & \multicolumn{9}{c|}{WER (\%)}                                                       \\ \cline{2-10} 
	& Advice & Child. & Comedy & Compet. & Docum. & Drama & Even. & News & {\bf Overall} \\ \hline
	--                         & 27.6   & 29.1   & 47.8    & 28.2    & 31.3   & 52.0  & 38.1  & 17.9 & {\bf 33.3}    \\ \hline
	SAT                         & 26.2   & 27.5   & 46.1    & 25.9    & 29.8   & 49.3  & 35.8  & 15.9 & {\bf 31.4}    \\ \hline
	LDaT                        & 25.8   & 27.8   & 45.1    & 25.7    & 28.9   & 47.7  & 33.5  & 15.7 & {\bf 30.6}    \\ \hline
	LDaT+SAT                    & 24.2   & 26.5   & 43.8    & 23.6    & 27.3   & 45.0  & 31.6  & 14.3 & {\bf 28.9}    \\ \hline
\end{tabular}	
\end{table*}%


\section{LDA Domain Analysis}
It is of interest to understand how the data is structured by the LDA model. Unfortunately ground 
truth labelling is only available for words and 8 different genres, and for both labels the quality 
is highly variable. One would suspect that the words themselves are less important, however acoustic 
attributes such as the presence of music or laughter may be very informative. Unfortunately such 
labels are not available. However one can still get some impression on the differences by looking at the
raw relationship to genres and differences between individual shows. 

\begin{figure}
	\centering
	\includegraphics[width=7cm]{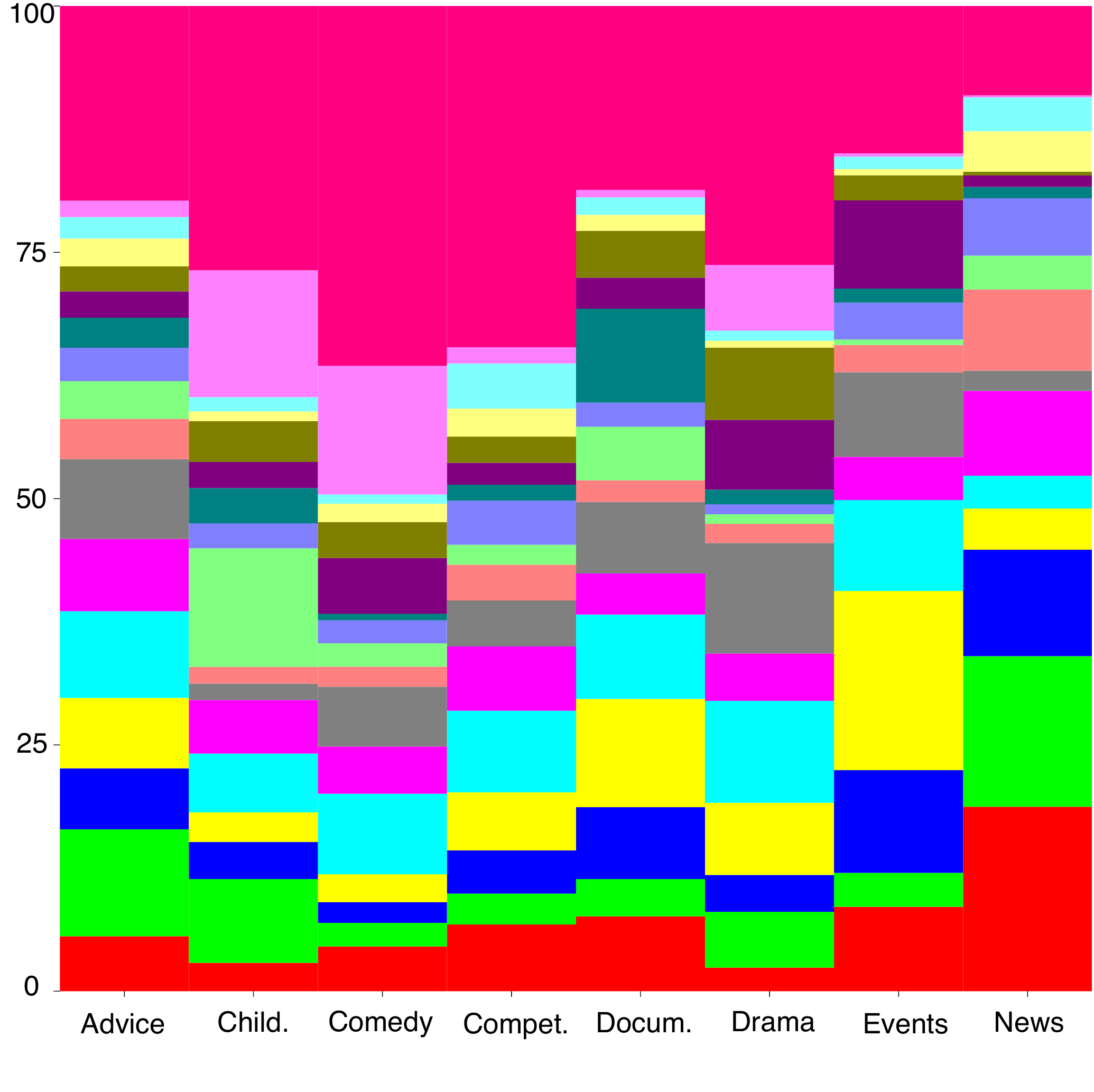}
	\caption{Distribution of data for the most important 
		16 LDA domains across genres}
	\label{fig:top16-lda}
\end{figure}

The domain assignment with the procedure outlined above, is visualised 
for the training data. The amount of data (time) assigned to an LDA domain is accumulated, for each 
of the 8 genres. Figure \ref{fig:top16-lda} shows the distribution of data for the most important 
16 LDA domains (based on duration), across genres. All remaining domains have been subsumed into 
one group at the top of the figure, for better illustration. In this and the following figure LDA 
domains are sorted by the amount of data overall. From the graph it is clear that different genres
exhibit significantly different LDA domain composition, with significant fine structure. Therefore 
such domain classification is very useful for genre classification. 

\begin{figure}
	\centering
	\includegraphics[width=7cm]{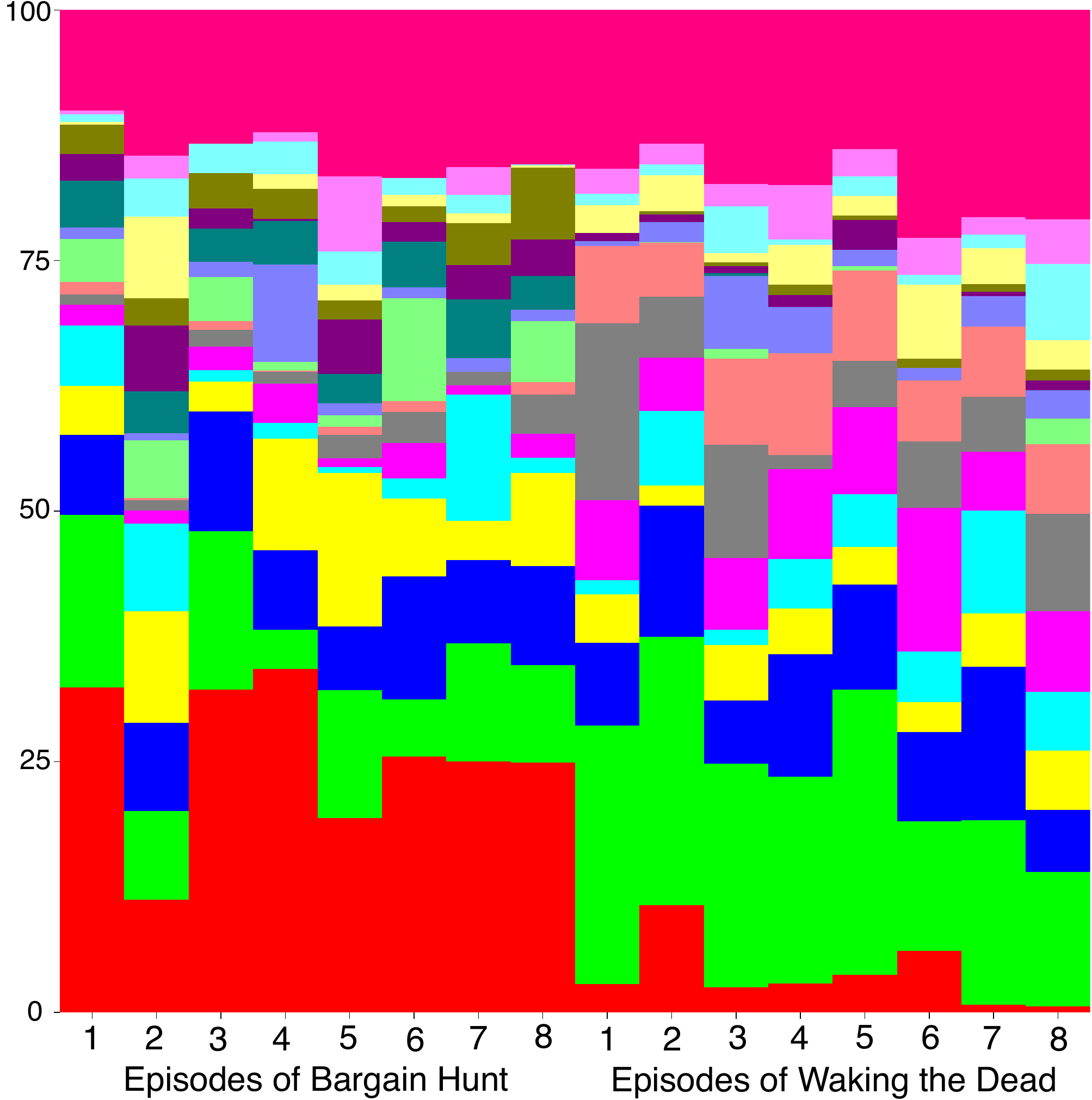}
	\caption{Within genre and between genre LDA domain distribution}
	\label{fig:show16-lda}
\end{figure}

One can also investigate how the LDA domain assignment varies within a genre, and between 
genres. In particular multiple episodes of shows are interesting in such analysis as one should 
expect high similarity due to similar programme structure. 
We obtained distributions for two sample programmes from two different genres. Figure 
\ref{fig:show16-lda} shows the LDA domain distribution for 8 episodes of \emph{Bargain Hunt} 
(competition), followed by a further 8 from \emph{Waking the Dead} (drama). 
Again 16+1 domains are displayed. 
One can observe that the distribution shows similarity within a genre 
(e.g., similarities of the red region on the lower left corner or the green area on the lower right 
corner). However between the two genres clear and systematic differences can be observed. 
One can further observe that more than 50\% of each show is typically described by the top 2 or 3 
LDA domains, and these differ in case of different genres but agree for the same 
programme within the genre. This indicates that individual shows are far more consistently 
described than the accumulated statistic allows to observe. 

\section{Conclusion}
\label{sec:conclusion}
This paper introduced a new method, latent-domain-aware training, to adapt Deep neural 
networks to new domains. The method employs acoustic Latent Dirichlet Allocation to identify 
acoustically distinctive data clusters. These so-called LDA domains are then encoded using one--hot 
encoding, and used to augment standard input features for DNNs in training and testing. We further 
introduced coherence data selection to improve classification quality, and presented results on a 
diverse set of BBC TV broadcasts, with 500h of training and 28h of testing data. Word Error Rate reduction of 13\% relative was achieved using the proposed adaptation method, compared to 
the baseline hybrid DNNs.

The proposed method lends itself to several future investigations. In the current LDA domain 
representation, each domain is described as a point on one of the axes of a high--dimensional 
space, where all have same distance from each other. Representing these points differently so that 
similar domains became closer in that space and verifying how that improves the performance can be 
an interesting problem to verify as a future work. Newer sets of features, better targeted to 
describe background acoustic characteristics \cite{saz2014slt}, could also provide an improvement.

\section{Data Access Statement}
\label{sec:data-access-statement}
The audio and subtitle data used for these experiments was distributed as part of the MGB Challenge 
(\url{mgb-challenge.org}) \cite{MGB} through a licence with the BBC. 
\bibliographystyle{IEEEbib}
\bibliography{refs}

\begin{thebibliography}{10}

\bibitem{Woodland97}
P.~C. Woodland, M.~J.~F. Gales, D.~Pye, and S.~J. Young,
\newblock ``Broadcast news transcription using {HTK},''
\newblock in {\em Proc. of ICASSP}, Munich, Germany, 1997.

\bibitem{Gauvain02}
J.~L. Gauvain, L.~Lamel, and G.~Adda,
\newblock ``The {LIMSI} broadcast news transcription system,''
\newblock {\em Speech Communication}, vol. 37, no. 1--2, pp. 89--108, 2002.

\bibitem{Gales06}
M.~J.~F. Gales, D.~Y. Kim, P.~C. Woodland, H.~Y. Chan, D.~Mrva, R.~Sinha, and
  S.~E. Tranter,
\newblock ``Progress in the {CU-HTK} broadcast news transcription system,''
\newblock {\em IEEE Trans. on Audio, Speech and Language Processing}, vol. 14,
  no. 5, pp. 1513--1525, 2006.

\bibitem{Lanchantin13}
P.~Lanchantin, P.~Bell, M.~Gales, T.~Hain, X.~Liu, Y.~Long, J.~Quinnell,
  S.~Renals, O.~Saz, and M.~Seigel,
\newblock ``Automatic transcription of multi--genre media archives,''
\newblock in {\em Proc. of SLAM}, Marseille, France, 2013.

\bibitem{doulatyis15}
M.~Doulaty, O.~Saz, and T.~Hain,
\newblock ``Data-selective transfer learning for multi-domain speech
  recognition,''
\newblock in {\em Proc. of Interspeech}, Dresden, Germany, 2015.

\bibitem{deng2015}
D.~Yu and L.~Deng,
\newblock {\em Automatic Speech Recognition: A Deep Learning Approach},
\newblock Springer-Verlag, London, UK, 2015.

\bibitem{Abrash95connectionistspeaker}
V.~Abrash, H.~Franco, A.~Sankar, and M.~Cohen,
\newblock ``Connectionist speaker normalization and adaptation,''
\newblock in {\em Proc. of EuroSpeech}, Madrid, Spain, 1995.

\bibitem{li2010comparison}
B.~Li and K.~C. Sim,
\newblock ``Comparison of discriminative input and output transformations for
  speaker adaptation in the hybrid nn/hmm systems,''
\newblock in {\em Proc. of Interspeech}, Makuhari, Japan, 2010.

\bibitem{gemello2007linear}
R.~Gemello, F.~Mana, S.~Scanzio, P.~Laface, and R.~De~Mori,
\newblock ``Linear hidden transformations for adaptation of hybrid {ANN/HMM}
  models,''
\newblock {\em Speech Communication}, vol. 49, no. 10, pp. 827--835, 2007.

\bibitem{stadermann2005two}
J.~Stadermann and G.~Rigoll,
\newblock ``Two-stage speaker adaptation of hybrid tied-posterior acoustic
  models.,''
\newblock in {\em Proc. of ICASSP}, Philadelphia, USA, 2005.

\bibitem{doddlipatla_is14}
R.~Doddipatla, M.~Hasan, and T.~Hain,
\newblock ``Speaker dependent bottleneck layer training for speaker adaptation
  in automatic speech recognition,''
\newblock in {\em Proc. of Interspeech}, Singapore, 2014.

\bibitem{dupont2000fast}
S.~Dupont and L.~Cheboub,
\newblock ``Fast speaker adaptation of artificial neural networks for automatic
  speech recognition,''
\newblock in {\em Proc. of ICASSP}, Istanbul, Turkey, 2000.

\bibitem{saon2013speaker}
G.~Saon, H.~Soltau, D.~Nahamoo, and M.~Picheny,
\newblock ``Speaker adaptation of neural network acoustic models using
  {i-Vectors},''
\newblock in {\em Proc. of ASRU}, Olomouc, Czech Republic, 2013.

\bibitem{liu2015}
Y.~Liu, P.~Karanasou, and T.~Hain,
\newblock ``An investigation into speaker informed {DNN} front-end for
  {LVCSR},''
\newblock in {\em Proc. of ICASSP}, Brisbane, Australia, 2015.

\bibitem{blei2003latent}
D.~M. Blei, A.~Y. Ng, and M.~I. Jordan,
\newblock ``{Latent Dirichlet Allocation},''
\newblock {\em Journal of Machine Learning Research}, vol. 3, pp. 993--1022,
  2003.

\bibitem{kim2009acoustic}
S.~Kim, S.~Narayanan, and S.~Sundaram,
\newblock ``Acoustic topic model for audio information retrieval,''
\newblock in {\em Proc. of WASPAA}, New Paltz NY, USA, 2009, pp. 37--40.

\bibitem{doulaty2015lda}
M.~Doulaty, O.~Saz, and T.~Hain,
\newblock ``Unsupervised domain discovery using latent dirichlet allocation for
  acoustic modelling in speech recognition,''
\newblock in {\em Proc. of Interspeech}, Dresden, Germany, 2015.

\bibitem{sivic2005discovering}
J.~Sivic, B.~C. Russell, A.~A. Efros, A.~Zisserman, and W.~T. Freeman,
\newblock ``Discovering objects and their location in images,''
\newblock in {\em Proc. of ICCV}, Beijing, China, 2005.

\bibitem{hu2009probabilistic}
D.~Hu and L.~K. Saul,
\newblock ``A probabilistic topic model for unsupervised learning of musical
  key-profiles.,''
\newblock in {\em Proc. of ISMIR}, Kobe, Japan, 2009.

\bibitem{griffiths2004finding}
T.~L. Griffiths and M.~Steyvers,
\newblock ``Finding scientific topics,''
\newblock {\em Proc. of National Academy of Sciences of the United States of
  America}, vol. 101, pp. 5228--5235, 2004.

\bibitem{gersho1992vector}
A.~Gersho and R.~M. Gray,
\newblock {\em Vector quantization and signal compression},
\newblock Springer Science \& Business Media, Berlin, Germany, 1992.

\bibitem{ni2015dataselection}
C.~Ni, C.~C. Leung, L.~Wang, N.~F. Chen, and B.~Ma,
\newblock ``Unsupervised data selection and word--morph mixed language model
  for tamil low-resource keyword search,''
\newblock in {\em Proc. of ICASSP}, Brisbane, Australia, 2015.

\bibitem{MGB}
P.~Bell, M.~J.~F. Gales, T.~Hain, J.~Kilgour, P.~Lanchantin, X.~Liu,
  A.~McParland, S.~Renals, O.~Saz, M.~Webster, and P.~Woodland,
\newblock ``The {MGB Challenge}: Evaluating multi-genre broadcast media
  recognition,''
\newblock in {\em Proc. of ASRU}, Arizona, USA, 2015.

\bibitem{MGB_ASR_SHEF}
O.~Saz, M.~Doulaty, S.~Deena, R.~Milner, R.~W.~M. Ng, M.~Hasan, Y.~Liu, and
  T.~Hain,
\newblock ``The 2015 {Sheffield} system for transcription of multi--genre
  broadcast media,''
\newblock in {\em Proc. of ASRU}, Arizona, USA, 2015.

\bibitem{ng2014iwslt}
R.~W.~M. Ng, M.~Doulaty, R.~Doddipatla, O.~Saz, M.~Hasan, T.~Hain, W.~Aziz,
  K.~Shaf, and L.~Specia,
\newblock ``The {USFD} spoken language translation system for {IWSLT} 2014,''
\newblock in {\em Proc. of IWSLT}, Lake Tahoe NV, USA, 2014.

\bibitem{Long13}
Y.~Long, M.~J.~F. Gales, P.~Lanchantin, X.~Liu, M.~S. Seigel, and P.~C.
  Woodland,
\newblock ``Improving lightly supervised training for broadcast
  transcriptions,''
\newblock in {\em Proc. of Interspeech}, Lyon, France, 2013.

\bibitem{hermansky1990perceptual}
H.~Hermansky,
\newblock ``Perceptual linear predictive ({PLP}) analysis of speech,''
\newblock {\em the Journal of the Acoustical Society of America}, vol. 87, no.
  4, pp. 1738--1752, 1990.

\bibitem{haeb1992linear}
R.~Haeb-Umbach and H.~Ney,
\newblock ``Linear discriminant analysis for improved large vocabulary
  continuous speech recognition,''
\newblock in {\em Proc. of ICASSP}, San Francisco, USA, 1992.

\bibitem{gales1998mllr}
M.~Gales,
\newblock ``Maximum likelihood linear transformations for {HMM}-based speech
  recognition,''
\newblock {\em Computer Speech \& Language}, vol. 12, no. 2, pp. 75 -- 98,
  1998.

\bibitem{anastasakos1996compact}
T.~Anastasakos, J.~McDonough, R.~Schwartz, and J.~Makhoul,
\newblock ``A compact model for speaker-adaptive training,''
\newblock in {\em Proc. of ICSLP}, Philadelphia, USA, 1996.

\bibitem{povey2008boosted}
D.~Povey, D.~Kanevsky, B.~Kingsbury, B.~Ramabhadran, G.~Saon, and
  K.~Visweswariah,
\newblock ``Boosted {MMI} for model and feature-space discriminative
  training,''
\newblock in {\em Proc. of ICASSP}, Las Vegas, USA, 2008.

\bibitem{hinton2006fast}
G.~E. Hinton, S.~Osindero, and Y.-W. Teh,
\newblock ``A fast learning algorithm for deep belief nets,''
\newblock {\em Neural Computation}, vol. 18, no. 7, pp. 1527--1554, 2006.

\bibitem{povey_2011_kaldi}
D.~Povey, A.~Ghoshal, G.~Boulianne, L.~Burget, O.~Glembek, N.~Goel,
  M.~Hannemann, P.~Motlicek, Y.~Qian, P.~Schwarz, J.~Silovsky, G.~Stemmer, and
  K.~Vesely,
\newblock ``The {Kaldi} speech recognition toolkit,''
\newblock in {\em Proc. of ASRU}, Hawaii, USA, 2011.

\bibitem{stolcke2002srilm}
A.~Stolcke,
\newblock ``Srilm-an extensible language modeling toolkit.,''
\newblock in {\em Proc. of Interspeech}, Denver, US, 2002.

\bibitem{rehurek_lrec}
R.~Rehurek and P.~Sojka,
\newblock ``Software framework for topic modelling with large corpora,''
\newblock in {\em Proc. of LREC}, Valletta, Malta, 2010.

\bibitem{saz2014slt}
O.~Saz, M.~Doulaty, and T.~Hain,
\newblock ``Background--tracking acoustic features for genre identification of
  broadcast shows,''
\newblock in {\em Proc. of SLT}, Lake Tahoe NV, USA, 2014.

\end{thebibliography}

\end{document}